\documentclass[runningheads]{llncs}

\usepackage{eccv}

\usepackage{eccvabbrv}

\usepackage{graphicx}
\usepackage{booktabs}
\usepackage{multirow}
\usepackage[accsupp]{axessibility}  

\usepackage{hyperref}

\usepackage{orcidlink}

\begin{document}
\newcommand{\Robert}[1]{\textcolor{red}{[Robert: #1]}}
\newcommand{\Reza}[1]{\textcolor{pink}{[RG: #1]}}
\newcommand{\Faizan}[1]{\textcolor{green}{[FS: #1]}}
\newcommand{\Enna}[1]{\textcolor{orange}{[ES: #1]}}
\newcommand{\fix}{\marginpar{FIX}}
\newcommand{\new}{\marginpar{NEW}}
\newcommand{\npar}[1]{\noindent\textbf{#1}}
\newcommand{\spar}[1]{\smallskip\noindent\textbf{#1}}
\newcommand{\mpar}[1]{\medskip\noindent\textbf{#1}}


\title{ESTANet: Efficient Online Error Detection in Procedural Videos via Prediction Inconsistency} 

\titlerunning{ESTANet}

\author{Shih-Po Lee\inst{1,2}\thanks{Work done as intern at Honda Research Institute, USA.}
\and
Reza Ghoddoosian\inst{1} \and Faizan Siddiqui\inst{1} \and Enna Sachdeva\inst{1} \and Behzad Dariush\inst{1}}
\authorrunning{S.-P. Lee et al.}

\institute{$^1$ Honda Research Institute, USA \qquad $^2$ Northeastern University}

\maketitle

\begin{abstract}
    An efficient and accurate system for detecting errors in procedural tasks is crucial for supporting human needs in daily life, as it can provide instant notifications and guide people to correct mistakes. In this work, we study real-time online error detection in procedural videos from a simple but overlooked perspective: the prediction behavior of action detectors themselves. Instead of designing complex architectures or specialized supervision, we observe that action detectors naturally exhibit different prediction characteristics depending on their sensitivity to input dynamics and temporal context. We therefore propose ESTANet (Error-Sensitive and Temporally-vArying Network), a lightweight framework that detects errors by exploiting inconsistencies among action predictions produced by a small set of action detectors. We construct standard and error-sensitive action detectors that behave similarly on correct executions but respond differently when errors occur. Meanwhile, detectors operating with different temporal contexts further amplify prediction inconsistencies when the procedure deviates from the intended sequence. During inference, we detect errors by aggregating mismatches between standard and error-sensitive predictions through majority voting to flag frames that contain errors. Extensive experiments on EgoPER, Assembly-101-O, and EPIC-Tent-O demonstrate that ESTANet achieves state-of-the-art performance in online error detection while maintaining real-time efficiency with a lightweight architecture. Our results highlight that leveraging the intrinsic properties of action detectors can yield a powerful and practical solution for online error detection without increasing architectural design complexity. Our code is available at: \url{https://github.com/robert80203/ESTANet}
\end{abstract}

\section{Introduction}
People have benefited greatly from advances in procedural video understanding across a wide range of problem domains and applications, including action detection \cite{Xu:ICCV19,Guo:ECCV22,Xu:NIPS21,Eun:CVPR20,Wang:ICCV21,Zhao:ECCV22,An:ICCV23,Wang:ICCV23,Pang:CVPR25}, segmentation \cite{Lu:CVPR24,Shen:CVPR24,Lu:ICCV25,Farha:CVPR19,Li:TPAMI20,Yi:BMVC21,Souri:PAMI21,Li:CVPR22,Liu:ICCV23}, video grounding \cite{Lu:CVPR25,Li:ECCV24,Mu:CVPR24,Dvornik:NIPS21,Dvornik:CVPR23,Ashutosh:NeurIPS23}, and error understanding \cite{Wang-Kwon:ICCV23,Ghoddoosian:ICCV23,Ding:arxiv23,Flaborea:CVPR24,Lee:CVPR24,Seminara:NeurIPS24,Huang:CVPR25, Patsch:ICCV25}. Imagine preparing lunch with an AI-assisted system that guides you through each step. If you make a mistake, the assistant provides instant feedback, enabling you to correct the error and continue seamlessly. For such assistance to be useful, error detection must be both real-time and online, operating only on past frames as people perform tasks sequentially. 
Meanwhile, these AI assistants are often deployed on wearable devices such as Microsoft HoloLens or Apple Vision Pro. They provide a natural first-person view and enable hands-free operation, but they also impose strict efficiency and latency constraints. Thus, \textit{development of a real-time and online error detection system for egocentric procedural videos} is crucial for enhancing both the user experience and task proficiency.

Recent studies have categorized errors in procedural videos into two types: execution errors and procedural errors \cite{Flaborea:CVPR24,Lee:CVPR24,Seminara:NeurIPS24,Huang:CVPR25}. Execution errors occur when actions interrupt the procedure, are not executed correctly, or are extra and not defined in the task. Examples include accidentally dropping a knife while preparing food, adding sugar instead of honey, or putting oats onto a tortilla which is not necessary for quesadilla preparation. In contrast, procedural errors involve repeating, omitting, or misordering existing steps. For instance, tightening all table legs before inserting a cross support violates the correct assembly sequence. Execution errors are well captured in datasets like EgoPER \cite{Lee:CVPR24} and CaptainCook4D \cite{Peddi:NeurIPS24}, while procedural errors are the focus of PREGO \cite{Flaborea:CVPR24}, derived from Assembly-101 \cite{Sener:CVPR22} and EPIC-Tent \cite{Jang:ICCV19}.


Existing methods tackle these errors either offline or online with only normal (error-free) videos during training. Offline methods \cite{Lee:CVPR24,Huang:CVPR25} require full videos and jointly perform temporal action segmentation and error detection. Online methods, in contrast, process current and past frames causally. Although recent online methods show strong benchmark performance, they suffer from key shortcomings:
(1) They only detect the first error in a video, limiting real-world applicability where multiple mistakes are common such as in EgoPER \cite{Lee:CVPR24} and CaptainCook4D \cite{Peddi:NeurIPS24}. (2) They mainly address procedural errors while overlooking nuanced execution errors. For example, PREGO \cite{Flaborea:CVPR24} relies on discrepancies between Large Language Model (LLM)-based anticipation and recognition labels, which hinders real-time performance and assumes correctness of prior predictions. DTGL \cite{Seminara:NIPS24} flags deviations from learned task graphs but fails to capture execution errors, where actions appear in sequence but are incorrectly performed. MistSense \cite{Patsch:ICCV25} requires hand poses as additional inputs and both error and error-free videos with annotations during training. 

We propose Error-Sensitive and Temporally-vArying Network (ESTANet) for real-time online error detection, which captures both procedural and execution errors by leveraging a simple yet effective principle: prediction inconsistencies naturally emerge when action detectors with different sensitivities and temporal contexts observe erroneous procedures. Our framework is an ensemble of two standard and two error-sensitive action detectors (four detectors in total) with the following components: (1) Each standard detector contains only an one-layer Gated Recurrent Unit (GRU),  followed by a linear classification head to predict action classes. (2) Each error-sensitive action detector contains a GRU, classification head, and Temporal-Aware Dynamic (TAD) module to produce inconsistent predictions, especially when execution errors occur. TAD generates dynamic weights and biases for the affine transformation applied to each frame feature as the input for the GRU (Fig. \ref{fig:tad_architecture}). The weights and biases are temporal-aware, input-dependent, and therefore, more sensitive to errors. (3) To enhance sensitivity to procedural errors, we introduce a temporally-varying design into our action detectors. Specifically, temporally-varying refers to training detectors with different temporal window sizes, such that each detector observes a distinct amount of temporal context and learns dependencies at different temporal scales. Therefore, when procedural errors occur, the detectors produce distinct predictions as they function based on different temporal context they have learned. However, since the space of possible temporal window sizes is combinatorially large, empirically training the models with all combinations to find the optimal one is time-consuming. To resolve it, we propose a robust strategy to efficiently determine the effective window sizes of each dataset. (4) During inference, we use the final action predictions to construct four predefined comparison pairs. For each frame, we examine whether the two predictions within each pair agree (i.e., predict the same action class). We then perform majority voting over the four pairwise agreement results to determine whether the frame contains an error. Note that we follow the same setting in PREGO \cite{Flaborea:CVPR24} and DTGL \cite{Seminara:NIPS24} to only train our model on error-free videos as it aligns with real-world scenario where normal videos are relatively easy to gather.


\section{Related Work}
\mpar{Online Action Detection.} 
Prior works \cite{Xu:ICCV19,Guo:ECCV22,Absil:book08,Wang:ICCV21,Zhao:ECCV22,An:ICCV23,Wang:ICCV23,Pang:CVPR25} have widely studied Online Action Detection (OAD), one of the popular directions in procedural video understanding. Given a video containing multiple actions, an OAD model identifies the actions taking place using only past and current frames. Specifically, miniROAD \cite{An:ICCV23} is built based on RNNs and adopts selected weights to only train the last frame of a given sequence of frames to alleviate the discrepancy between training and inference. The design is simple yet effective for addressing OAD. A recent work, CMeRT \cite{Pang:CVPR25}, further alleviates training-inference discrepancy in previous OAD methods due to imbalanced context exposure in long- and short-term memory. It adopts a context-enhanced encoder using near-past context to learn consistent short-term representations, and a memory-refined decoder uses near-future context with learned short-term representations to detect actions. We construct our proposed method based on miniROAD.

\mpar{Error Detection in Procedural Videos.} The research community has recently shown growing interest in error detection, which aims to detect the occurrence of or localize an action that changes the action sequence of a procedure or should not occur in a procedure. Specifically, among the offline methods, EgoPED \cite{Lee:CVPR24} determines the execution errors by thresholding the similarities between the predicted action features and input frames features. AMNAR \cite{Huang:CVPR24} generates action features conditioned on executed actions and follows the same strategy as in EgoPED for error detection. On the other hand, among the online methods, PREGO \cite{Flaborea:CVPR24} detects procedural errors by the difference in prediction generated by an action detector and LLM, with its strong reasoning capability to anticipate the next action given past actions. DTGL \cite{Seminara:NIPS24} learns the task graph from the training videos and detects an action as a procedural error if its preconditions in the learned task graph are not in the observed actions. MistSense \cite{Patsch:ICCV25} trains a mistake detection module with hand poses and videos in an end-to-end and fully-supervised manner and a LLM to explain the detected errors.

\begin{figure}[t]
  \centering
  \includegraphics[width=\linewidth, trim=0 10cm 0 0, clip]{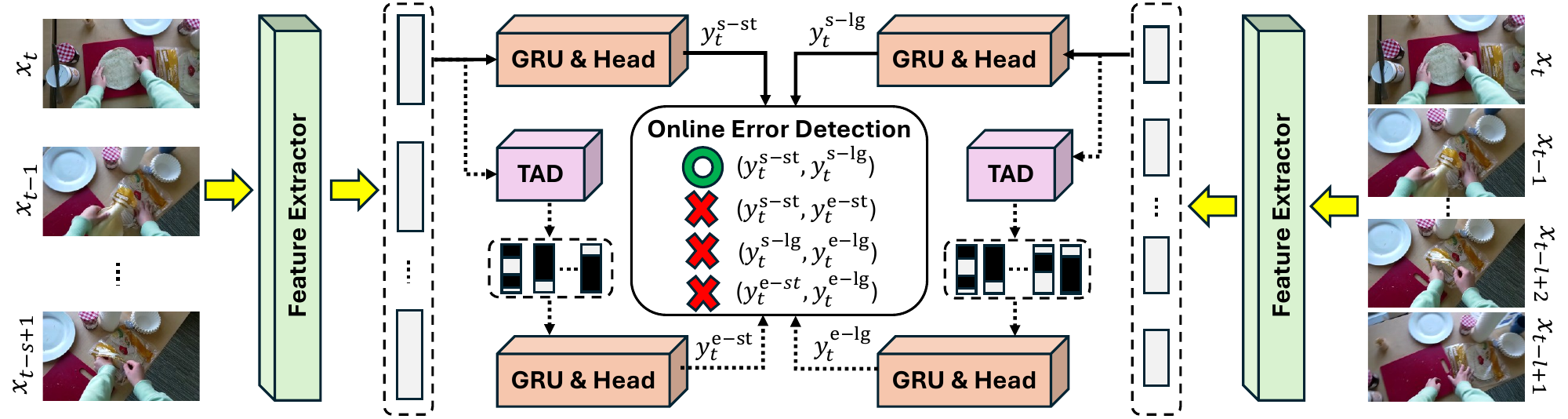}
  \caption{\footnotesize{The pipeline of our ESTANet. At time $t$, it produces action predictions $y^{\text{s-st}}_t$ and $y^{\text{s-lg}}_t$ by standard detectors and $y^{\text{e-st}}_t$ and $y^{\text{e-lg}}_t$ by error-sensitive detectors trained on small and large window sizes ($s$ and $l$ frames). The final error prediction is obtained by majority voting over the four agreement pairs based on action predictions.
  }}
  \label{fig:architecture}
\end{figure}

\section{Proposed Method}
\label{proposed_method}

\subsection{Problem Setting and Framework Overview}
Given a frame $x_t$ at time $t$ in a video with length $T$ and past frame $(x_0, \ldots, x_{t-1})$, we aim to predict $z_t \in \{0, 1\}$, where $z_t=1$ indicates an error occurs in frame $x_t$, otherwise $z_t=0$. This procedure starts from the first frame and continues until the end of the video. Note that all training videos are error-free and only the ground-truth frames-wise action classes $(\bar{y}_0, \bar{y}_1, \ldots, \bar{y}_T)$ are available, where $\bar{y}_t \in \{0, 1, \ldots, P\}$, $P$ is the number of actions, and $\bar{y}_t=0$ denotes background.

Our proposed ESTANet is illustrated in Fig. \ref{fig:architecture}.
Given an input frame at time $t$, ESTA produces four action predictions. The two standard detectors (upper region of Fig.~\ref{fig:architecture}) output $y^{\textrm{s-st}}_t$ and $y^{\textrm{s-lg}}_t$, while the two error-sensitive detectors with the Temporal Aware Dynamic (TAD) module (lower region of Fig.~\ref{fig:architecture}) output $y^{\textrm{e-st}}_t$ and $y^{\textrm{e-lg}}_t$. Here, s and e denote standard and error-sensitive detectors, while st and lg denote models trained with small and large temporal window sizes $s$ and $l$.
During inference, we combine four predictions into four comparison pairs based on the rationale of capturing procedural and execution errors (Section \ref{subsec:online_error_detection}), and an error is detected if at least three of the pairs mismatch (middle of Fig. \ref{fig:architecture}).  In the following sections, we will describe 1) the design of standard and error-sensitive action detectors, 2) the robust selection strategy to find window sizes for temporally-varying attribute in action detectors, 3) the training losses for our ESTANet, and 4) online error detection during inference.

\subsection{Standard and Error-Sensitive Action Detector}
\label{subsec:ad}
We construct standard action detectors to produce stable action predictions and the sensitive ones to predict inconsistent actions where their inconsistencies capture especially execution errors. We begin with the forward pass of standard action detectors trained with window size $s$ (the upper-left region in Fig. \ref{fig:architecture}). The standard action detector processes a sequence of $s$ frames $(x_{t-s+1},\cdots,x_{t-1},x_t)$ ending at time $t$ through a GRU and a MLP as the classification head to predict the action probability $a^{\textrm{s-st}}_t \in \mathbb{R}^{P+1}$ at time $t$, followed by argmax to obtain predicted action class $y^{\textrm{s-st}}_t$. We use this forward pass with $l$ frames $(x_{t-l+1},\cdots,x_{t-1},x_t)$ to obtain action probability $a^{\textrm{s-lg}}_t \in \mathbb{R}^{P+1}$ and class $y^{\textrm{s-lg}}_t$. 

\begin{figure}[t]
  \centering
  \includegraphics[width=1.0\linewidth, trim=15 11.5cm 5cm 5, clip]{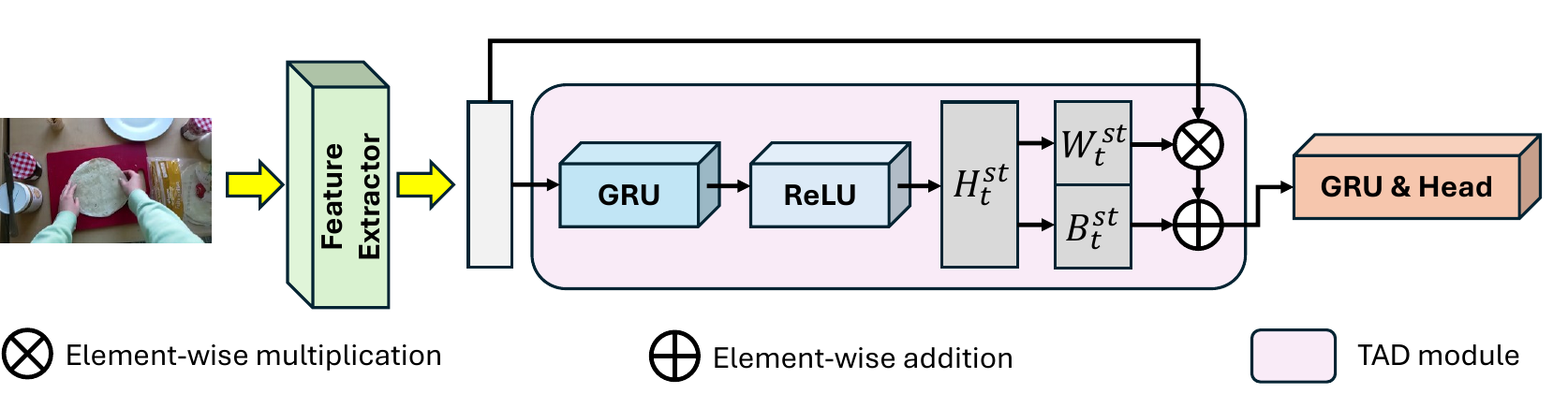}
  \caption{\footnotesize{Pipeline of the TAD module with window size $s$. The TAD module with window size $l$ uses the same pipeline.}}
  \vspace{-5pt}
  \label{fig:tad_architecture}
\end{figure}

Next, we construct error-sensitive action detectors that are sensitive to execution errors which are spatiotemporally different from their correct actions (the bottom-left region in  Fig. \ref{fig:architecture}). We propose a TAD module to convert frame features into the representations that are more sensitive to erroneous frames (see Fig. \ref{fig:tad_architecture}). Take window size $s$ as an example. TAD uses a GRU and ReLU to generate temporal-aware and input-dependent vector $H^{\text{st}}_t \in \mathbb{R}^{2D}$ at time $t$, and subsequently divides $H^{\text{st}}_t$ into $W^{\text{st}}_t$ as the weight vector and $B^\text{st}_t$ as the bias vector, where $W^{\text{st}}_t, B^{\text{st}}_t \in \mathbb{R}^{D}$. Since the weight and bias vectors are input-dependent, the GRU and classification head trained on such frame features after the affine transformation are more input-sensitive, especially when out-of-distribution data (e.g., execution errors) occur, resulting in inconsistent action predictions. Overall, the error-sensitive action detectors produce action probabilities $a^{\textrm{e-st}}_t$ and $a^{\textrm{e-lg}}_t$ with window size $s$ and $l$, where $a^{\textrm{e-st}}_t, a^{\textrm{e-lg}}_t\in \mathbb{R}^{P+1}$ at time $t$, followed by argmax to obtain predicted action class $y^{\textrm{e-st}}_t$ and $y^{\textrm{e-lg}}_t$, respectively.


\subsection{Temporally-Varying Attribute}
\label{subsec:ta}
We introduce a robust strategy to efficiently instantiate a temporally-varying attribute within our ESTANet for procedural error detection. This attribute is realized by training action detectors with different temporal window sizes, inducing distinct temporal receptive fields. As a result, each detector captures dependencies at different temporal scales. When procedural errors occur, the inconsistency between short- and long-context predictions is amplified, providing a discriminative signal for error detection.

We construct our action detectors based on GRUs (Section \ref{subsec:ad}), which possess several properties: (1) They encode inductive biases toward temporal invariance and locality through their recurrent (Markovian) state transition structure \cite{battaglia:arxiv2018}. (2) They exhibit distinct prediction behaviors at inference time when trained with different temporal window sizes (i.e., clip durations), as demonstrated in MiniROAD \cite{An:ICCV23}. Therefore, we leverage action detectors trained with varying temporal window sizes to capture inconsistencies in their predictions, which naturally arise from this design when past or current actions deviate from the correct execution sequence (e.g., procedural errors). 

However, a key question arises: \textbf{what temporal window sizes are required to effectively enable temporally-varying behavior in action detectors, particularly when procedural errors occur?} One straightforward solution is to heuristically train action detectors using different pairs of temporal window sizes and select the optimal configuration by jointly varying $s$ and $l$.  Nonetheless, this process is time-consuming and lacks a guiding strategy, requiring the same exhaustive search to be repeated for each dataset. As a result, we propose a robust strategy to automatically select the temporal window sizes $s$ and $l$ for each dataset. Our strategy (1) enables distinct prediction behaviors under procedural errors, and (2) substantially reduces the heuristic search required to determine effective window sizes.

\begin{figure}[t]
  \centering
  \includegraphics[width=1.0\linewidth, trim=1 11cm 0 0, clip]{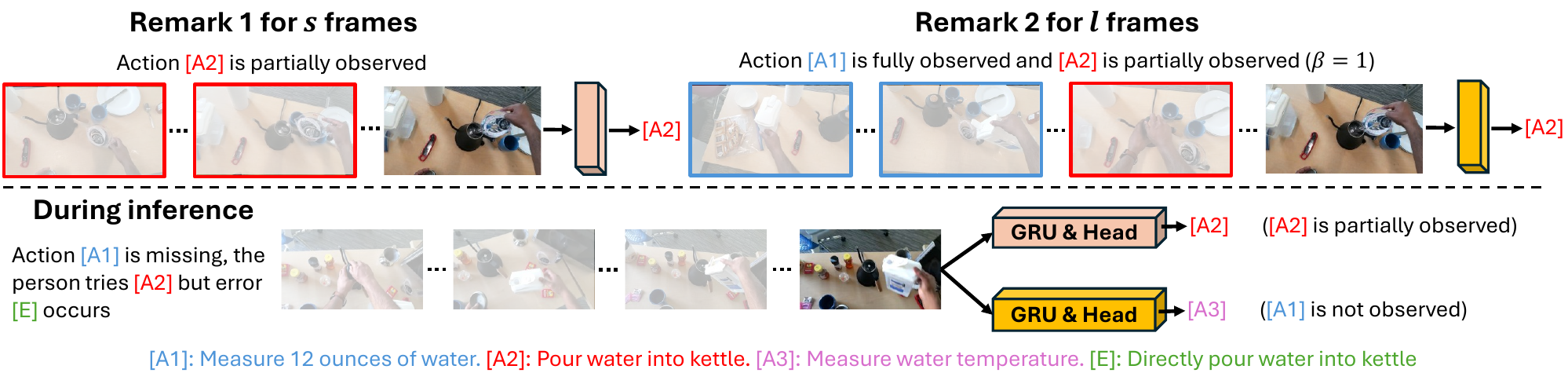}
  \caption{\footnotesize{Illustration of temporally-varying attribute. The bottom region demonstrates an example where errors occur (missing Action [A1] and doing Error [E]).}}
  \label{fig:principle}
\end{figure}

We begin by determining $s$ (the upper-left region of Fig.~\ref{fig:principle}). Given a dataset, we compute the average duration of each action and select the one with the shortest mean duration as $s$. This choice ensures that the action detector primarily observes partial past context within the current action when making predictions. Next, for $l$ (the upper-right region of Fig.~\ref{fig:principle}), we automatically determine a window size such that at the start time of approximately $\theta$\% of the actions in the dataset, the $l$-frame window covers $\beta$ complete preceding actions. This strategy enables the action detector to predict the current action using complete information from preceding actions. Consequently, when procedural errors occur, the detector trained with window size $l$ exhibits prediction behaviors distinct from the one trained with window size $s$, as it operates with complete preceding actions (see Fig.~\ref{fig:qualitative_missing_steps} in the Experiments section). On the other hand, we set $\theta = 50$, which provides a dataset-agnostic and robust criterion that balances covering too many versus too few preceding actions, as real-world videos often interleave short and long actions. It avoids extreme cases, for example, when a small number of long actions would force a $\theta = 100$ threshold to span an entire action, thereby causing excessive overlap with relatively short actions and leading to imbalanced context of preceding actions. We highlight two key insights below.

\spar{Remark 1.} The choice of $s$ encourages that the action detector has learned to predict the action class by only seeing partial ongoing action, making it utilize short-range dependency and current frame.

\spar{Remark 2.} The design of $l$ enables the action detector depending on learned long-range dependency, specifically, the context spanning complete preceding and partial ongoing actions, when predicting the current action.

The strategy ensures our action detectors are temporally-varying regarding the dataset in an automatic fashion, and produce the inconsistencies in predictions when procedural errors occur. For example, the bottom region of Fig. \ref{fig:principle} shows that the detector trained with window size $s$ predicts action \textcolor{red}{A2}, whereas the detector trained with window size $l$ predicts action \textcolor{purple}{A3}, since the context of preceding \textcolor{blue}{A1} is absent, thereby altering its prediction behavior. 

\subsection{Training Losses}
ESTANet is trained on batches of frame sequences, each independently and randomly sampled from the training videos. Concretely, each batch consists of training pairs $(v,t)$ from a random time $t$ and video $v$. Next, we describe the training losses applied to the standard and error-sensitive action detectors.

We adopt the same cross-entropy loss as in MiniROAD \cite{An:ICCV23}. The action loss is computed for the action probability of the frame only at time $t$ while observing $s$ or $l$ prior frames. Additionally, we utilize a smoothing loss that minimizes the difference between two consecutive predictions in time. The two losses for each sample in the batch are formulated below.
\begin{equation}
    \mathcal{L}_{\text{CE}}(t) = - \sum_{p \in \mathcal{A}} \log(p_{t,\bar{y}_{t}}), \ \mathcal{L}_{\text{S}}(t) = \sum_{p \in \mathcal{A}}\frac{1}{P+1}  \sum_{i=0}^{P} \left( \log  p_{t,i}  - \log  p_{t-1,i} \right)^2
\end{equation}

where $p$ denotes action probability, $\mathcal{A} = \{a^{\text{s-st}}, a^{\text{s-lg}}, a^{\text{e-st}}, a^{\text{e-lg}}\}$, and $\bar{y}_t$ is the ground-truth action label. For each sample in the batch, the classification loss $\mathcal{L}_{\textrm{cls}}(t)=\mathcal{L}_{\textrm{CE}}(t)+\mathcal{L}_{\textrm{S}}(t)$ is defined as the sum of the two losses for all detectors.

\subsection{Online Error Detection during Inference}
\label{subsec:online_error_detection}

At inference time, we process the video frame by frame from $t=1$ to $t=T$ to obtain predictions from the action detectors. At each time $t$, we 1) apply a causal mode filter, which only covers past frames to replace the current action with the most frequent action within a window size to smooth the predictions, and 2) perform error detection via action prediction inconsistency. Note that, due to the use of causal mode filter, the error flag may be delayed by a second.


\spar{Detecting Execution Errors.} We form two pairs: ($y^{\text{s-st}}_t$, $y^{\text{e-st}}_t$) and ($y^{\text{s-lg}}_t$, $y^{\text{e-lg}}_t$), where $y^{\text{s-st}}_t$ and $y^{\text{s-lg}}_t$ provide stable action predictions on errors and $y^{\text{e-st}}_t$ and $y^{\text{e-lg}}_t$ provide inconsistent action predictions as they are error-sensitive. Inconsistencies in the two pairs are valuable for capturing execution errors, since subtle variations often differentiate correct actions from errors. 

\spar{Detecting Procedural Errors.} We form another two pairs: ($y^{\text{s-st}}_t$, $y^{\text{s-lg}}_t$) and ($y^{\text{e-st}}_t$, $y^{\text{e-lg}}_t$), generated by the action detectors trained with small and large window sizes ($s$ and $l$ frames). Inconsistencies observed in the two pairs imply that the predicted action does not follow any correct action sequence and thus, facilitate the detection of procedural errors. 

Finally, in a real-world scenario, multiple types of errors can occur in one video and interactively influence each other. To finally determine whether a frame contains an error or not, we use majority voting on the mismatches for those four pairs. We flag a frame as an error ($z_t=1$) if at least three mismatches are detected, otherwise as a correct action ($z_t=0$).

\section{Experiments}

\subsection{Experimental Setup}
\spar{Dataset.}
We evaluate our proposed method on EgoPER \cite{Lee:CVPR24}, Assembly-101-O \cite{Flaborea:CVPR24}, and EPIC-Tent-O \cite{Flaborea:CVPR24}. EgoPER consists of 5 tasks (\textit{tea}, \textit{quesadilla}, \textit{oatmeal}, \textit{pinwheel}, and \textit{coffee}) with 386 egocentric videos and contains both execution and procedural errors. Each video may contain both types of errors. We use the given training/test split for evaluation. For Assembly-101-O and EPIC-Tent-O, which consist of egocentric videos with procedural errors in assembly domain, we follow the same training/test split as in PREGO \cite{Flaborea:CVPR24} and DTGL \cite{Seminara:NIPS24}. Note that the procedure in each test video for Assembly-101-O and EPIC-Tent-O stops once an error occurs, meaning that only the last action is flagged as an error.

\spar{Evaluation Metrics.}
For EgoPER, we report the metric used in \cite{Farha:CVPR19}, the segment-wise F1 score under three overlap thresholds (10\%, 25\%, and 50\%) based on normal and erroneous segments, denoted as F1@10, F1@25, and F1@50. They consider both localization and classification performance of error detection, especially for execution errors. For Assembly-101-O and EPIC-Tent-O, we report the same action-wise F1 scores as in DTGL for two complementary classification cases: erroneous $\rightarrow$ correct (E-F1) and correct $\rightarrow$ erroneous (C-F1). Their average yields Avg-F1. All F1 scores are computed based on predicted actions.


\spar{Baselines.}
We mainly compare our proposed method with three state-of-the-art online error detection methods, MistSense \cite{Patsch:ICCV25}, PREGO \cite{Flaborea:CVPR24} and DTGL \cite{Seminara:NIPS24} with direct optimization which performs the best in general. For EgoPER, we train PREGO with Qwen 2.5-14B-Instruct \cite{AnYang:Arxiv24} as its LLM and DTGL from scratch with their public codes and evaluate their performance. We only report the performance of MistSense on EPIC-Tent-O as there is no public code available. For Assembly-101-O and EPIC-Tent-O, we additionally include other two baselines, MSGI \cite{Sohn:ICLR20} and $\textrm{MSG}^2$ \cite{Jang:ICLR23} reported in DTGL. 

\spar{Implementation Details.}
For EgoPER, we use TimeSFormer \cite{Bertasius:ICML21}, pre-trained on Ego4D \cite{Grauman:CVPR22}, to extract frame-level features using 4 past frames as input at 10 frames per second (FPS). For Assembly-101-O and EPIC-Tent-O, we use the frame features generated by PREGO at 30 and 60 FPS, respectively. Every GRU has one layer and every classification head has one fully-connected layer. We set $\beta=2$ and $\theta=50\%$ across all the datasets. We train the model using AdamW optimizer with learning rate 0.0001 and weight decay 0.05 for 20 epochs. Note that for Assembly-101-O and EPIC-Tent-O, we generate a sequence of action segments based on $y^{\text{s-st}}$ and flag the action as an error if any frame within the segment is detected as an error. In addition, to simplify the setting where all datasets use FPS = $10$, we uniformly sample one frame for every $3$ frames in Assembly-101-O and every $6$ frames in EPIC-Tent-O during training and inference, respectively. We set the window size of mode filter to $30$. See our supplementary material for more details. We train and evaluate all models using Intel Xeon(R) Silver 4210 CPU@2.20GHz and NVIDIA RTX A6000.

\begin{table}[t]
    \footnotesize
    \caption{\footnotesize{Online error detection performance on EgoPER, Assembly-101-O, and EPIC-Tent-O. $\dagger$ indicates the model is trained with both normal and erroneous videos.}}
    \vspace{-2.5pt}
    \setlength{\tabcolsep}{3.2pt} 
    \centering
    \begin{tabular}{lccccccccc}
        \toprule
        \multirow{2}{*}{Method} & \multicolumn{3}{c}{EgoPER} & \multicolumn{3}{c}{Assembly-101-O} & \multicolumn{3}{c}{EPIC-Tent-O} \\   
         & F1@10 & F1@25 & F1@50 & Avg-F1 & E-F1 & C-F1 & Avg-F1 & E-F1 & C-F1 \\
        \cmidrule(r){1-1} \cmidrule(r){2-4} \cmidrule(r){5-7} \cmidrule(r){8-10} 
        MSGI \cite{Sohn:ICLR20} & - & - & - & 33.1 & 43.5 & 22.7 & 44.5 & 22.0 &  66.9 \\
        $\textrm{MSG}^2$ \cite{Jang:ICLR23} & - & - & - & 46.2 & 33.2 & 59.1  & 45.2 &  22.9 & 67.5 \\
        PREGO \cite{Flaborea:CVPR24} & 41.2 & 28.1 & 12.2 & 32.5 & 41.8 & 23.1 & 29.4 & 17.2 & 41.6 \\
        DTGL \cite{Seminara:NIPS24} & 39.6 & 33.8 & 20.9 & 53.5 & 28.1 & \textbf{78.9} & 46.5 & 23.7 & \textbf{69.3} \\
        MistSense$\dagger$ \cite{Patsch:ICCV25} & - & - & - & - & - & - & 59.8 & 89.7 & 29.8 \\

        ESTANet (Ours) & \textbf{47.2} & \textbf{37.8} & \textbf{21.6} &  \textbf{59.6}  & \textbf{49.6} & 69.5 & \textbf{70.2} & \textbf{92.4} &   48.0 \\
        
        \bottomrule
    \end{tabular}
    \label{tab:error_detection}
\end{table}

\subsection{Experimental Results}
\spar{Online Error Detection.}
Table \ref{tab:error_detection} compares ESTANet with existing methods on EgoPER, Assembly-101-O, and EPIC-Tent-O. ESTANet consistently outperforms all baselines. On EgoPER, ESTANet achieves 47.2\%, 37.8\%, and 21.6\% at F1@10, F1@25, and F1@50, respectively, compared with DTGL, which obtains 39.6\%, 33.8\%, and 20.9\%. On Assembly-101-O and EPIC-Tent-O, ESTANet further surpasses DTGL and MistSense by 6.1\% and 10.4\% on Avg-F1, respectively. We observe that DTGL exhibits a conservative bias, frequently predicting actions as correct and thus under-detecting errors. In contrast, MistSense tends to over-predict errors, resulting in excessive false positives. Benefiting from our TAD module and temporally-varying attribute, ESTANet achieves a more balanced prediction behavior, obtaining the highest E-F1 scores (49.6\% and 92.4\%) while maintaining competitive C-F1 scores (69.5\% and 48.0\%) on Assembly-101-O and EPIC-Tent-O, respectively. A detailed breakdown is provided in the supplementary material.




\begin{table}[t]
    \footnotesize
    \caption{\footnotesize{Complexity and inference speed analysis in terms of frame per second (FPS) and number of parameters (Size).}}
    \vspace{-2.5pt}
    \setlength{\tabcolsep}{8pt} 
    \centering
    \begin{tabular}{cccccc}
        \toprule
        \multirow{2}{*}{Method} & \multicolumn{2}{c}{Feature Extractor} & \multicolumn{2}{c}{Error Detection} & \multicolumn{1}{c}{Combined} \\
        & Size & FPS & Size & FPS & FPS \\
        \cmidrule(r){1-1} \cmidrule(r){2-3} \cmidrule(r){4-5} \cmidrule(l){6-6}
        PREGO \cite{Flaborea:CVPR24} & 121.4M & 38.2 & $\sim$ 14B & 1.38 & 1.33 \\
        DTGL \cite{Seminara:NIPS24} & 121.4M & 38.2 & 5.72M & 1180 & 37.0 \\
        MistSense \cite{Patsch:ICCV25} & 307.0M & 33.0 & $\sim$ 7B & 2.6 & 2.5 \\ 
        ESTANet (Ours) & 121.4M & 38.2 & 18.8M & 67.5 &  24.4 \\

        \bottomrule
    \end{tabular}
    \label{tab:inference_speed}
\end{table}

\begin{figure}[t]
  \centering
  \begin{subfigure}[t]{0.49\textwidth}
    \centering
    \includegraphics[width=\linewidth]{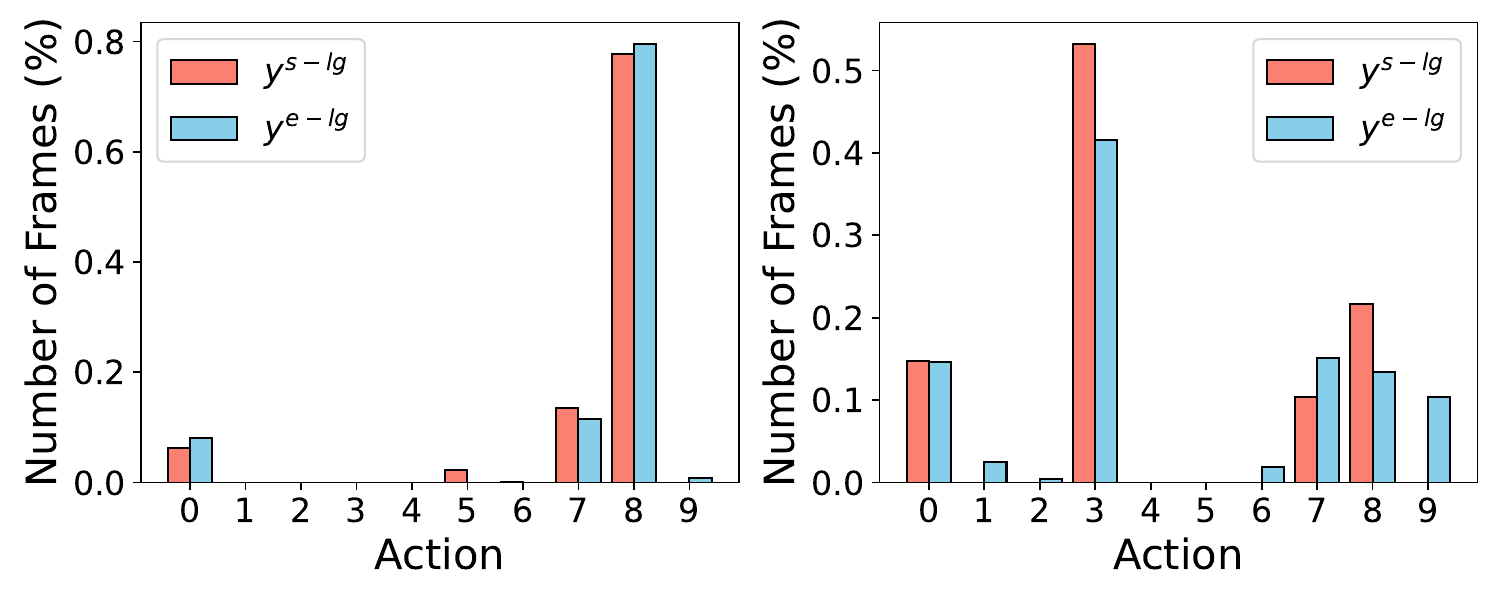} 
    \caption{\footnotesize{Correct action 8 (left): Add honey into tea. Erroneous action (right): Add sugar into tea instead.}}
    \label{fig:tea_act2_action_error_count}
  \end{subfigure}\hfill
  \begin{subfigure}[t]{0.49\textwidth}
    \centering
    \includegraphics[width=\linewidth]{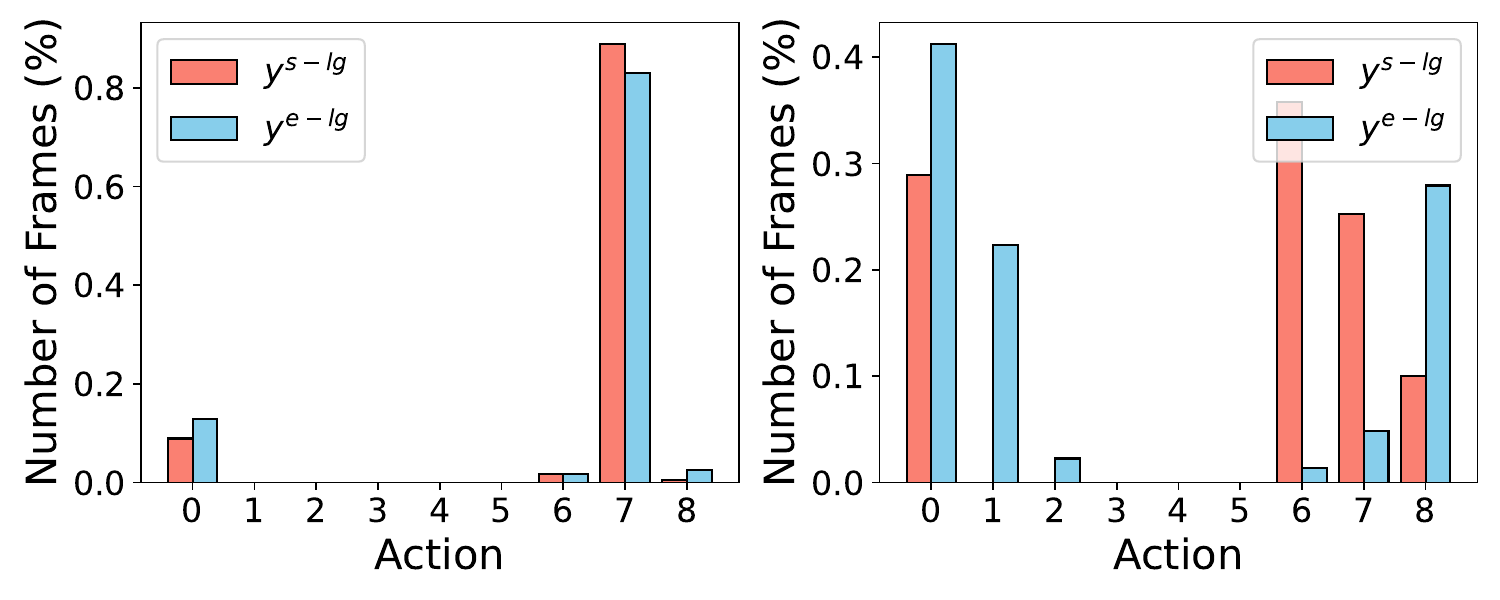}
    \caption{\footnotesize{Correct action 7 (left): Slice tortilla using knife. Erroneous action (right): Slice tortilla with hands.}}
    \label{fig:que_act7_action_error_count}
  \end{subfigure}

  \caption{\footnotesize{Histograms of frame-wise predicted actions $y^{\text{s-lg}}$ (red) and $y^{\text{e-lg}}$ (blue) for correct actions (left) and their corresponding execution errors (right) in tea (a) and quesadilla (b) from EgoPER. The x-axis denotes action categories, and the y-axis represents the percentage of frames predicted as each action.}}
  \label{fig:action_error_count}
\end{figure}

\mpar{Prediction Inconsistency Analysis.}
Fig. \ref{fig:action_error_count} presents the histograms of frame-wise predictions from the standard and error-sensitive detectors ($y^{\text{s-lg}}$ and $y^{\text{e-lg}}$) for two correct actions and their corresponding execution errors in EgoPER. In Fig. \ref{fig:action_error_count}(a) and (b) (left), both detectors exhibit similar prediction distributions for correct actions. Their predictions are highly concentrated on the corresponding ground-truth actions (action 8 for \textit{tea} and action 7 for \textit{quesadilla}), indicating consistent and stable behavior on normal actions. In contrast, the right panels show that the error-sensitive detector produces more diverse prediction patterns for erroneous actions. For example, in Fig. \ref{fig:action_error_count}(a) (right), $y^{\text{e-lg}}$ spreads across eight different action categories, whereas $y^{\text{s-lg}}$ distributes over only four categories for the error “Add sugar into tea instead.” Similarly, in Fig. \ref{fig:action_error_count}(b) (right), the two detectors exhibit noticeably different prediction distributions for the error “Slice tortilla with hands.” These results demonstrate that ESTANet amplifies prediction inconsistency on errors while maintaining consistency for correct actions.

\begin{table}[t]
    \footnotesize
    \caption{\footnotesize{The ablation for our TAD module}}
    \vspace{-2.5pt}
    \setlength{\tabcolsep}{3.2pt} 
    \centering
    \begin{tabular}{lccccccccc}
        \toprule
        \multirow{2}{*}{Method}  & 
        \multicolumn{3}{c}{EgoPER} & \multicolumn{3}{c}{Assembly-101-O} & \multicolumn{3}{c}{EPIC-Tent-O} \\   
         & F1@10 & F1@25 & F1@50 & Avg-F1 & E-F1 & C-F1 & Avg-F1 & E-F1 & C-F1 \\
        \cmidrule(r){1-1} \cmidrule(r){2-4} \cmidrule(r){5-7} \cmidrule(r){8-10} 
         w/o TAD & 35.3 & 22.5 & 7.5 & 55.1 & 45.3  & 64.9  & 69.3 & 93.1 & 45.5  \\
         w/ TAD & 47.2 & 37.8 & 21.6 &  59.6  & 49.6 & 69.5 & 70.2 & 92.4 & 48.0 \\
        \bottomrule
    \end{tabular}
    \label{tab:ablation_tad}
\end{table}

\begin{figure}[t]
  \centering
  \begin{subfigure}[t]{0.48\textwidth}
    \centering
    \includegraphics[width=\linewidth, trim=0 8cm 8cm 0cm, clip]{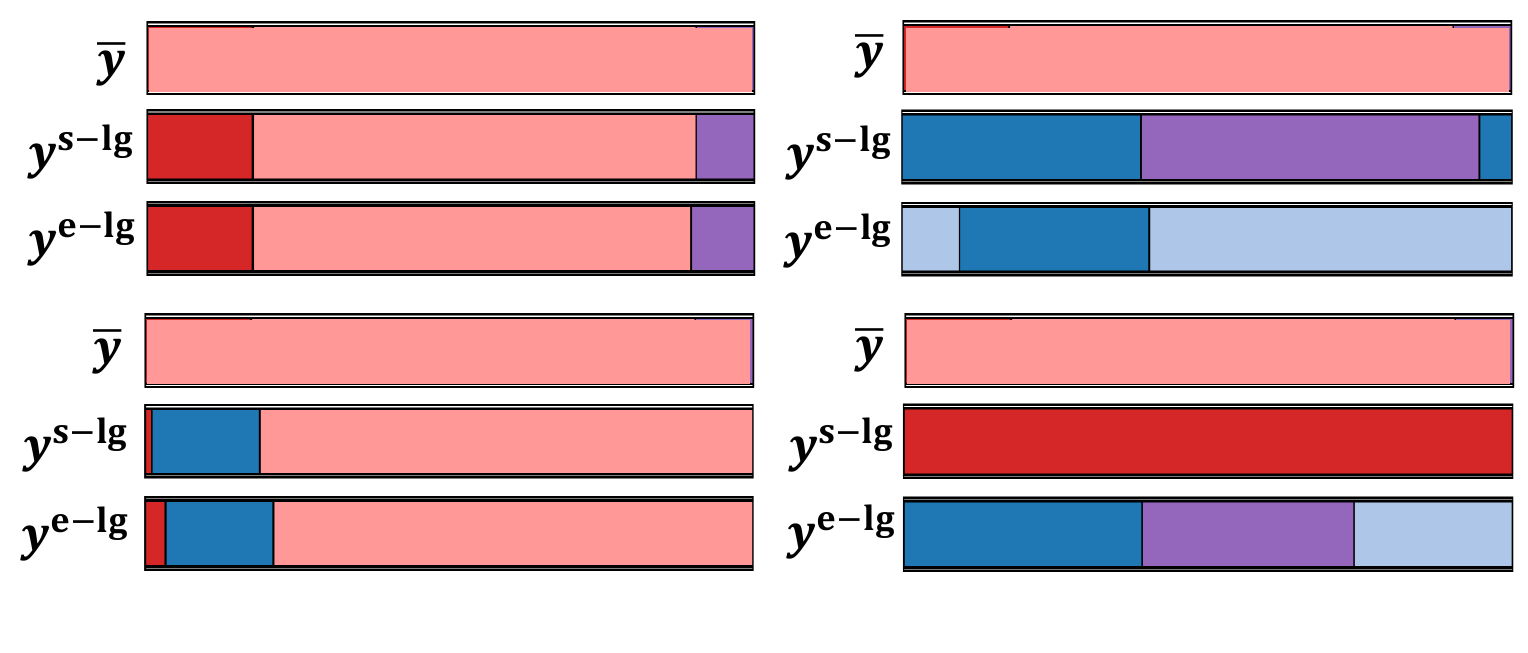} 
    \caption{\footnotesize{Correct action (left): Slice tortilla using knife. Erroneous action (right): Slice tortilla with hands instead of knife.}}
    \label{fig:slice_with_hand}
  \end{subfigure}\hfill
  \begin{subfigure}[t]{0.48\textwidth}
    \centering
    \includegraphics[width=\linewidth, trim=0 8cm 8cm 0cm, clip]{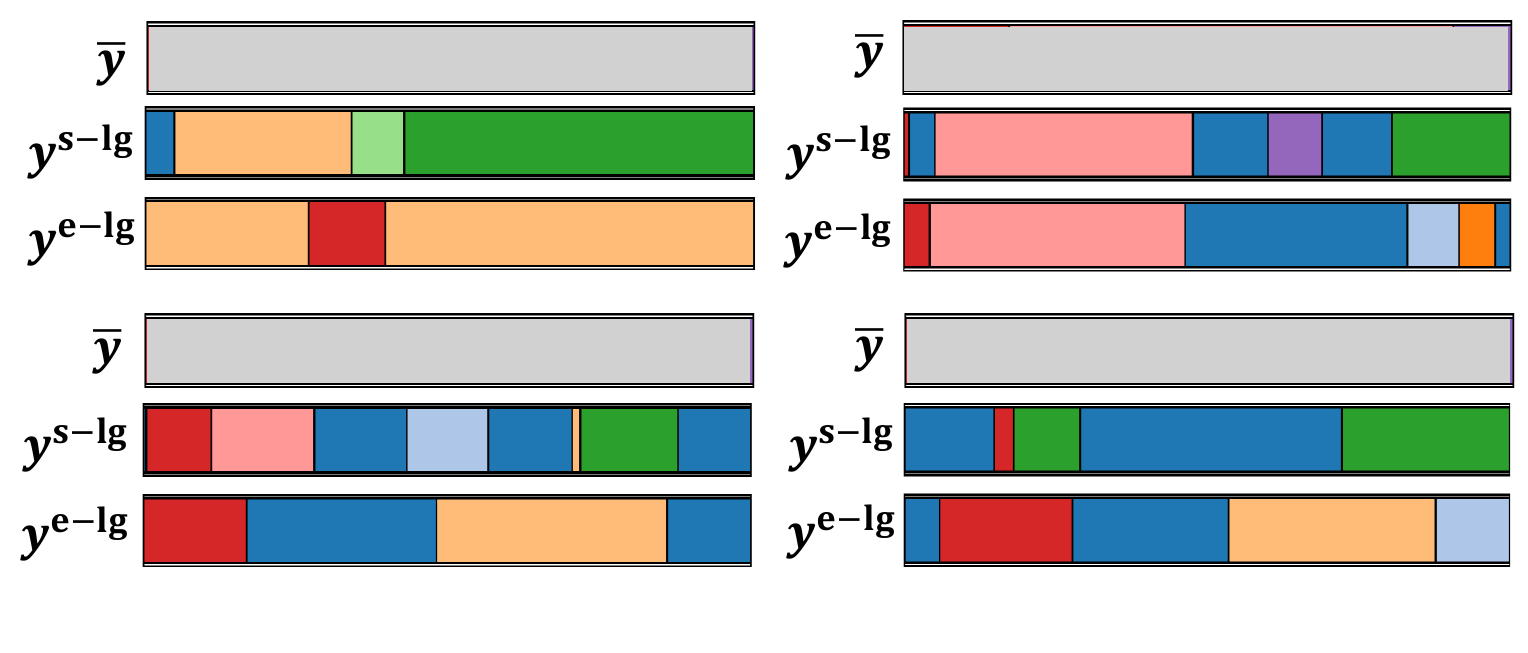}
    \caption{\footnotesize{All four figures denote the same unseen error: Put mug into microwave for 5 seconds.}}
    \label{fig:put_mug_into_microwave}
  \end{subfigure}

  \caption{\footnotesize{Each row in a sub-figure, from top to bottom shows frame-wise ground-truth action classes $\bar{y}$, $y^{\text{s-lg}}$ and  $y^{\text{e-lg}}$ on \textit{quesadilla} (a) and \textit{tea} (b) of EgoPER. Each color represents an action class.}}
  \label{fig:abl_tad}
\end{figure}

\mpar{Complexity Analysis.} Table \ref{tab:inference_speed} shows the comparison between different methods in terms of throughput and model size. Note that the combined setting better reflects real-world scenarios as it considers both frame processing and error detection. ESTANet achieves real-time processing (24.4 FPS) in the combined setting, and outperforms PREGO and MistSense, which attain 1.33 FPS with a LLM and 2.5 FPS with a heavier feature extractor and LLM, respectively. On the other hand, DTGL pre-computes the task graph, enabling no-delay error detection through pre-condition matching and only requiring a lightweight action detector (5.72M). Although DTGL obtains a higher FPS with fewer parameters than ESTANet, the latter surpasses the former in F1 scores across three datasets while maintaining real-time processing speed (24.4 FPS).

\mpar{Ablation for TAD Module.}
We quantitatively and qualitatively analyze our TAD module. Table \ref{tab:ablation_tad} shows that without our TAD module, the F1@50 scores drastically drops from 21.6\% to 7.5\%. The result indicate that TAD module effectively increases the error sensitivity in action detectors to capture execution errors. In addition, TAD module also improves the performance on both Assembly-101-O and EPIC-Tent-O with relatively small extent (1\% to 4\%) since they only contain procedural errors. On the other hand, we visualize the qualitative results on correct and erroneous actions. The figures in the left column of Fig. \ref{fig:abl_tad}(a) correspond to the correct actions from two videos and $y^{\text{s-lg}}$ and $y^{\text{e-lg}}$ are similar. In contrast, the right column refers to the erroneous action. 
Note that $y^{\text{s-lg}}$ shows relatively consistent predictions within the segment while $y^{\text{e-lg}}$ shows more diverse predictions, interleaved with different actions. On the other hand, the four figures in Fig. \ref{fig:abl_tad} (b) represent the same error (unseen action) from different videos. Both $y^{\text{s-lg}}$ and $y^{\text{e-lg}}$ show inconsistent action predictions but with different patterns, showing that our TAD module introduces a distinct prediction pattern that is helpful for capturing execution errors.


\mpar{Ablation for Temporally-varying Attribute.}
We study the effectiveness of our robust strategy for enabling temporally-varying attribute. We use the same $s$ for all methods as it provides limited temporal information for action detection and compare our proposed method with two other strategies for choosing $l$: (1) the duration of $2^{\text{th}}$ shortest action and (2) the duration of longest action. The former contains more partial context of an ongoing action, similar to what window size $s$ covers, whereas the latter one contains context of one or more complete preceding actions.

\begin{table}[t]
    \footnotesize
    \caption{\footnotesize{The ablation for choosing window size $l$ regarding different strategies.}}
    \vspace{-2.5pt}
    \setlength{\tabcolsep}{3pt} 
    \centering
    \begin{tabular}{cccccccccc}
        \toprule
        \multirow{2}{*}{Strategy} & \multicolumn{1}{c}{EgoPER} & \multicolumn{4}{c}{Assembly-101-O} & \multicolumn{4}{c}{EPIC-Tent-O} \\  
          &  F1@50 & Avg-F1 & E-F1 & C-F1 & $l$ & Avg-F1 & E-F1 & C-F1 & $l$ \\
        \cmidrule(r){1-1} \cmidrule(r){2-2} \cmidrule(r){3-6} \cmidrule(r){7-10}
          $2^{\textrm{th}}$ shortest Act. & 15.9 &  47.0 & 11.9 & \textbf{82.2} & 54 & 52.1 & 15.7 & \textbf{88.6} & 123 \\
         longest Act. & 15.2  & 45.3 & 16.3 & 74.3 & 633 & 47.0 & 12.5 & 81.5 & 551  \\
         $\beta=1$ (Ours) & 21.0 & \textbf{60.0} & \textbf{55.6} & 64.4 & 168 & 66.6 & \textbf{93.2} & 40.0 & 132  \\
         $\beta=2$ (Ours) & \textbf{21.6} & 59.6 & 49.6 & 69.5 & 400 & \textbf{70.2} & 92.4 & 48.0 & 440 \\
         $\beta=3$ (Ours) & 19.8 & 59.8 & 52.5 & 67.0 & 656 & 65.6 & 91.2 & 40.0 & 868 \\
        \bottomrule
    \end{tabular}
    \vspace{-2.5pt}
    \label{tab:ablation_stlg}
\end{table}

\begin{table}[t]
    \footnotesize
    \caption{\footnotesize{The ablation for different thresholds of our robust strategy for temporally-varying attribute ($\beta=2$).}}
    \vspace{-2.5pt}
    \setlength{\tabcolsep}{3pt} 
    \centering
    \begin{tabular}{cccccccccc}
        \toprule  
        \multirow{2}{*}{$\theta$(\%)}  & \multicolumn{1}{c}{EgoPER} & \multicolumn{4}{c}{Assembly-101-O} & \multicolumn{4}{c}{EPIC-Tent-O} \\  
         & F1@50 & Avg-F1 & E-F1 & C-F1 & $l$ & Avg-F1 & E-F1 & C-F1 & $l$ \\
        \cmidrule(r){1-1} \cmidrule(r){2-2} \cmidrule(r){3-6} \cmidrule(r){7-10}
          20 & 15.6 & 42.8 & 11.4 & 74.2 & 204 & 48.7 & 12.6 & \textbf{84.9} & 208 \\
          50 & \textbf{21.6} & \textbf{59.6} & \textbf{49.6} & 69.5 & 400 & \textbf{70.2} & \textbf{92.4} & 48.0 & 440 \\
          80 & 17.4 & 44.3 & 14.2 & \textbf{74.4} & 756 & 45.2 & 10.3 & 80.1 & 996 \\
        \bottomrule
    \end{tabular}
    \vspace{-2.5pt}
    \label{tab:ablation_threshold}
\end{table}

Table \ref{tab:ablation_stlg} shows that our proposed method with different $\beta$ consistently outperforms other strategies, specifically achieving approximately 5\%, 15\%, and 14\% higher on F1@50 for EgoPER, Avg-F1 for Assembly-101-O and EPIC-Tent-O with $\beta=2$, respectively. Note that small and large values for $l$ by our method can all effectively enable temporally-varying attribute in action detectors. Finding such values is non-trivial and our proposed method efficiently finds them. For example, for EPIC-Tent-O, $2^{\text{th}}$ shortest action ($l=123$) achieves 52.1\% on Avg-F1 while our method ($\beta=2$, $l=132$) achieves 66.6\% only with 9-frame difference. On the other hand, for Assembly-101-O, longest action ($l=633$) obtains 45.3\% on Avg-F1 whereas our method ($\beta=3$, $l=656$) obtains 59.8\%. The consistent improvement with different $l$ values by various $\beta$ demonstrates that our method can efficiently enable the temporally-varying attribute in our framework to capture procedural errors.


Finally, we ablate our threshold design ($\theta = 50$) for choosing the window size $l$, as shown in Table \ref{tab:ablation_threshold}. Using $\theta = 20$ results in insufficient coverage of complete preceding actions. In contrast, $\theta = 80$ enforces extensive coverage, overly including complete preceding actions. Both extremes lead to imbalanced temporal context modeling, where detectors are trained with disproportionate numbers of preceding actions, and weaken error sensitivity.
In comparison, $\theta = 50$ provides a balanced context of preceding actions, yielding the best performance of 21.6\% F1@50 on EgoPER, 59.6\% Avg-F1 on Assembly-101-O, and 70.2\% Avg-F1 on EPIC-Tent-O.

\begin{figure}[t]
  \centering
  \includegraphics[width=1.0\linewidth, trim=20 5 5 12cm, clip]{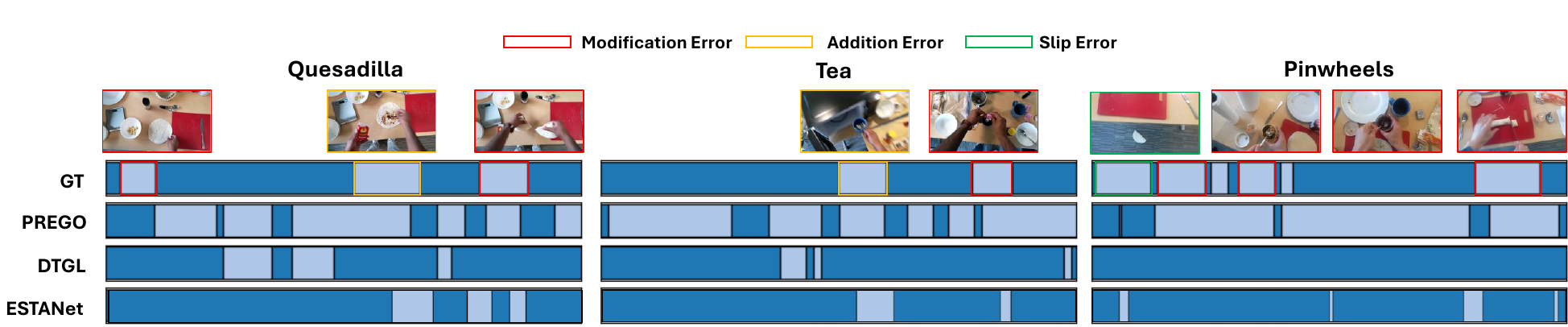}
  \caption{\footnotesize{Qualitative visualization of online error detection on EgoPER. Each row from top to bottom shows specific erroneous frames, GT error detection, and error detection predicted by PREGO, DTGL, and ESTANet.}}
  \label{fig:qualitative}
\end{figure}

\begin{figure}[t]
  \centering
  \includegraphics[width=1.0\linewidth, trim=0 15.5cm 10cm 0, clip]{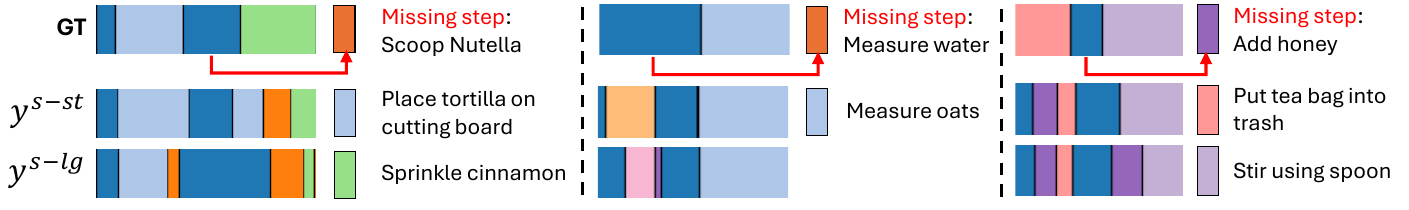}
  \caption{\footnotesize{Qualitative visualization of frame-wise predictions when procedural errors (missing steps) occur on EgoPER.}}
  \label{fig:qualitative_missing_steps}
\end{figure}

\mpar{Qualitative Analysis.} We visualize the online error detection results generated by different models in Fig. \ref{fig:qualitative}. Specifically, PREGO produces many false positive segments and DTGL barely detects the locations of the errors. Our ESTANet can detect the locations of execution errors, including various types defined by EgoPER. For instance, ESTANet can detect (1) a modification error, ``put tortilla on the table instead of the cutting board'' (marked in red) in task quesadilla, (2) an addition error, ``put mug into microwave'' (marked in yellow) in task tea, and (3) an slip error, ``drop tortilla on floor'' (marked in green) in task pinwheels. 

On the other hand, Fig. \ref{fig:qualitative_missing_steps} presents qualitative results of our action predictions ($y^{\text{s-st}}$ and $y^{\text{s-lg}}$) under procedural errors caused by missing steps. When the execution deviates from the correct action sequence, the detectors in our ESTANet produce distinct prediction patterns. Particularly, in the middle region of Fig. \ref{fig:qualitative_missing_steps}, $y^{\text{s-st}}$ and $y^{\text{s-lg}}$ generate different predictions for the missing step (marked in orange and pink). In the rightmost region, $y^{\text{s-lg}}$ hallucinates the omitted step (marked in dark purple), as it expects the complete procedural structure, whereas $y^{\text{s-st}}$ focuses primarily on the currently observed actions (marked in light purple). The results demonstrate that introducing varying temporal receptive fields can amplify prediction inconsistencies when procedural errors occur, substantiating the effectiveness of our temporally-varying attribute and robust strategy. See our supplementary material for more qualitative results.

\begin{figure}[t]
  \centering
  \includegraphics[width=1.0\linewidth, trim=0 14cm 6cm 0, clip]{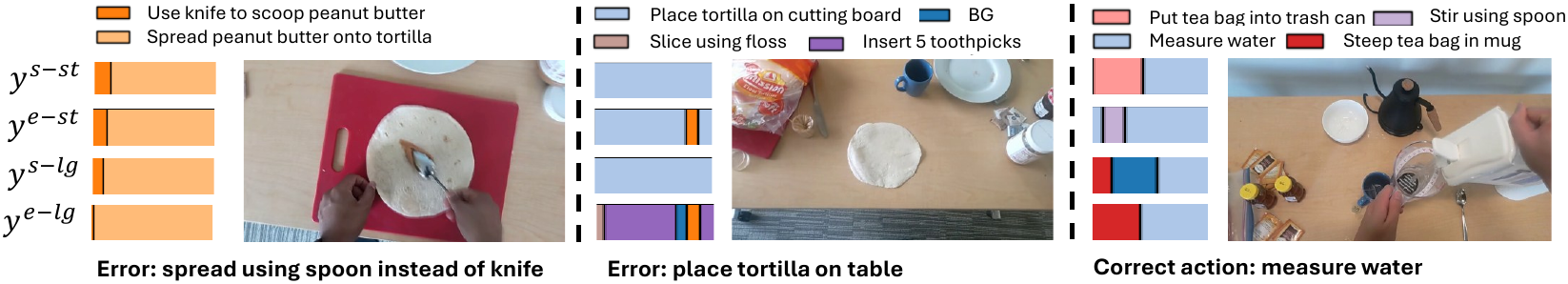}
  \caption{\footnotesize{Qualitative visualization of faliure cases on EgoPER.}}
  \vspace{-2.5pt}
  \label{fig:failure_case}
\end{figure}

\section{Limitations}
In this section, we analyze the failure cases of ESTANet. First, when the discrepancy between the correct action and the error is subtle (leftmost region in Fig. \ref{fig:failure_case}), all detectors produce similar predictions. This suggests that the current feature representation lacks sufficient spatial information. 
Second, the detectors exhibit strong temporal inductive bias, particularly at the beginning of a video where preceding context is absent (middle region in Fig. \ref{fig:failure_case}). In such cases, the models tend to predict the original correct action even when an obvious execution error occurs, indicating over-reliance on learned temporal priors. Finally, limited model capacity may lead to false positive predictions. For instance, in the rightmost region of Fig. \ref{fig:failure_case}, the person performs “measuring water” after “place tea bag in mug.” Due to the long background interval between actions, the short-context detectors produce inconsistent predictions ($y^{\text{s-st}}$ and $y^{\text{e-st}}$), while the long-context detectors predict the same wrong action “steep tea bag in mug” ($y^{\text{s-lg}}$ and $y^{\text{e-lg}}$). These observations highlight the trade-off between learning strong temporal inductive biases, achieving accurate action recognition, and maintaining error sensitivity. 

\section{Conclusions}
In this paper, we propose Error-Sensitive and Temporally-vArying Network (ESTANet) for real-time online error detection in procedural videos. ESTANet detects errors by utilizing prediction inconsistencies naturally exhibited by action detectors under different sensitivities and temporal contexts. Specifically, we construct standard and error-sensitive action detectors to capture execution errors through differences in prediction stability, and enable temporally-varying attributes by training detectors with different temporal window sizes to reveal procedural deviations. We further introduce a robust strategy for efficiently determining effective small and large temporal windows for each dataset. Extensive experiments on three datasets demonstrate that our lightweight framework effectively detects complex errors in real-world videos while maintaining real-time performance.

\bibliographystyle{splncs04}
\bibliography{biblio_bank/ehsan,biblio_bank/learning,biblio_bank/vision,biblio_bank/detection,biblio_bank/segmentation,biblio_bank/math,biblio_bank/procedurelearning,biblio_bank/action_understanding,biblio_bank/nlp,biblio_bank/anomaly_detection,biblio_bank/vision_language,biblio_bank/deeplearning,biblio_bank/error_understanding,biblio_bank/societal_impact,biblio_bank/vlm-llm}
\end{document}